\documentclass[letter, 10pt, conference]{ieeeconf}
\IEEEoverridecommandlockouts
\usepackage{graphicx}
\usepackage{amsmath}
\usepackage{amsfonts}
\usepackage{subcaption}
\usepackage{booktabs}
\usepackage{multirow}
\usepackage{url}
\usepackage{hyperref}
\usepackage{cleveref}
\usepackage{comment}
\usepackage[acronyms]{glossaries}
\usepackage{xcolor}

\crefname{section}{Sec.}{Secs.}
\crefname{table}{Tab.}{Tabs.}
\crefname{figure}{Fig.}{Figs.}
\crefname{equation}{Eq.}{Eqs.}

\title{\LARGE \bf
Fixed External Cameras as Common Prior Maps for Active 3D Scene Graph Generation
}

\author{Giorgia Modi$^{12*}$, Davide Buoso$^{2}$, Giuseppe Averta$^{2}$, Daniele De Martini$^{1}$ \\ 
$^{1}$Mobile Robotics Group (MRG), University of Oxford, UK \\
$^{2}$Visual and Multimodal Applied Learning Lab (VANDAL), Politecnico di Torino, Italy
\thanks{*Correspondent author: \texttt{giorgiamodi@gmail.com}}
\thanks{This work was supported by the EPSRC Impact Acceleration Account (IAA), and the UKRI InnovateUK project `A general-purpose configuration management system for highly configurable systems validated in automated logistics'. 
}
}

\begin{document}

\maketitle

\begin{abstract}

Commonly available prior information, such as BIM models, floor plans, and remote sensing images, can provide valuable geometric and semantic context for autonomous robotic systems. In this paper, we treat observations from fixed external RGB cameras as \emph{Common Prior Maps} (CPMs): wide-field views of the environment that initialize a semantic and geometric scene prior before any robot motion begins. We present an RGB-only framework for active, incremental 3D scene graph (3DSG) generation that seamlessly fuses observations from both onboard robot cameras and fixed external cameras within a single hardware-agnostic pipeline. By relying solely on RGB observations processed by a feed-forward 3D reconstruction model, the system treats all cameras - onboard or external - identically, requiring no hardware modifications. A graph-based active semantic exploration framework then directly leverages the partial scene graph to guide the robot toward regions of high semantic uncertainty, progressively completing and refining the prior. Experiments demonstrate that bootstrapping the scene graph with even a single external camera increases initial object recall by up to $\textbf{+79\%}$, and that the richer context of the prior significantly improves the efficiency of subsequent active exploration. 
\end{abstract}

\section{Introduction}
Modern robotic systems navigating unstructured human environments – such as homes, hospitals and offices - heavily depend upon rich contextual visual scene understanding to perceive, decide, and act autonomously~\cite{ni2023deep}. Traditional perception pipelines build either low-level dense metric maps that lack explicit semantic content, or produce object-centric outputs that fail to encode relational structure~\cite{xu2017iterative}. A structured representation that jointly captures \emph{which} entities are present, \emph{where} they exist in 3D space, and \emph{how} they relate to one another is therefore essential for reliable robot autonomy. 3D scene graphs (3DSGs) have emerged as a powerful representation for this purpose~\cite{Li2024survey,bae2023survey3Dscenegraphs}. In a 3DSG, nodes correspond to object instances annotated with semantic labels and metric locations, while edges encode spatial and functional relationships between them. This compact and queryable structure supports downstream tasks including language and task planning~\cite{rana2023sayplan}, navigation~\cite{werby2024hierarchical}, and manipulation~\cite{jiang2024roboexp}. However, constructing such representations robustly and efficiently in real-world environments remains challenging, especially when prior knowledge about the scene is only partially available.

This motivates the use of \emph{Common Prior Maps} (CPMs), which have long been available in real-world deployments and offer rich geometric and semantic priors. We argue that observations from fixed external cameras, already deployed as surveillance or monitoring infrastructure in many buildings, can serve as a valuable and underutilized form of CPM for robot scene understanding.

To exploit this prior, the perception pipeline must seamlessly incorporate external camera observations, which are typically available as RGB images. We therefore present a framework for active 3D scene graph generation from RGB inputs alone. By removing the dependency on depth sensors and relying instead on feed-forward 3D reconstruction~\cite{keetha2025mapanything}, the system processes all RGB inputs - onboard and external - through a unified pipeline. Within this framework, the 3DSG becomes the semantic and geometric representation extracted from external-camera observations, capturing the dominant layout, objects, and spatial relationships of the environment.

Our contributions are:
\begin{enumerate}
\item an RGB-only pipeline for 3D scene graph generation that enables seamless integration of fixed external cameras;
\item an active semantic exploration loop that uses the scene graph to select informative viewpoints;
\item experimental evidence that camera-infrastructure priors substantially improve scene graph completeness and exploration efficiency.
\end{enumerate}

\section{Related Work}
\label{sec:related_work}

\textbf{3D Scene Graphs (3DSGs).}
Scene graphs couple object semantics with relational context, addressing ambiguities in purely object-centric systems~\cite{xu2017iterative}. Grounding this abstraction in 3D enables navigation, manipulation, and language-guided planning~\cite{catalano20253D}. Systems such as Hydra~\cite{hughes2022hydra} and SceneGraphFusion~\cite{wu2021scenegraphfusion} build hierarchical representations from RGB-D SLAM, while ConceptGraphs~\cite{gu2024conceptgraphs} introduces open-vocabulary semantics via foundation models, but still assumes access to depths and camera poses. To the best of our knowledge, our system is the first to enable 3DSG generation in an RGB-only setting.

\textbf{Active Perception.}
Active perception—controlling sensing to maximize information gain—has long been studied via Next-Best-View (NBV) planning \cite{bajcsy2018revisiting}. Geometric frontier-based methods \cite{yamauchi1997frontier} and density-based approaches \cite{border2024surface} efficiently expand coverage but ignore semantics. Recent work incorporates semantic reasoning: Ding et al.~\cite{ding2025sea} use semantic scene completion to guide exploration toward uncertain regions, while Chen et al.~\cite{chen2025understanding} combine semantic and geometric uncertainty via 3D Gaussian Splatting.  ASP \cite{tang2025active} further enables LLM-based reasoning over scene graphs, turning them into active exploration tools. However, prior work on ASP has not considered settings with external camera inputs, nor RGB-only perception.

\textbf{External sensing.}
While robotic perception is typically egocentric, fixed external cameras provide complementary global context. Prior work shows their value for visual servoing \cite{robinson2026robot} and VLM-guided navigation \cite{buoso2025select2plan}, but their integration into 3DSGs remains unexplored. 

\section{Methodology}
\label{sec:methodology}

\begin{figure*}[t]
\centering
\includegraphics[width=0.95\textwidth]{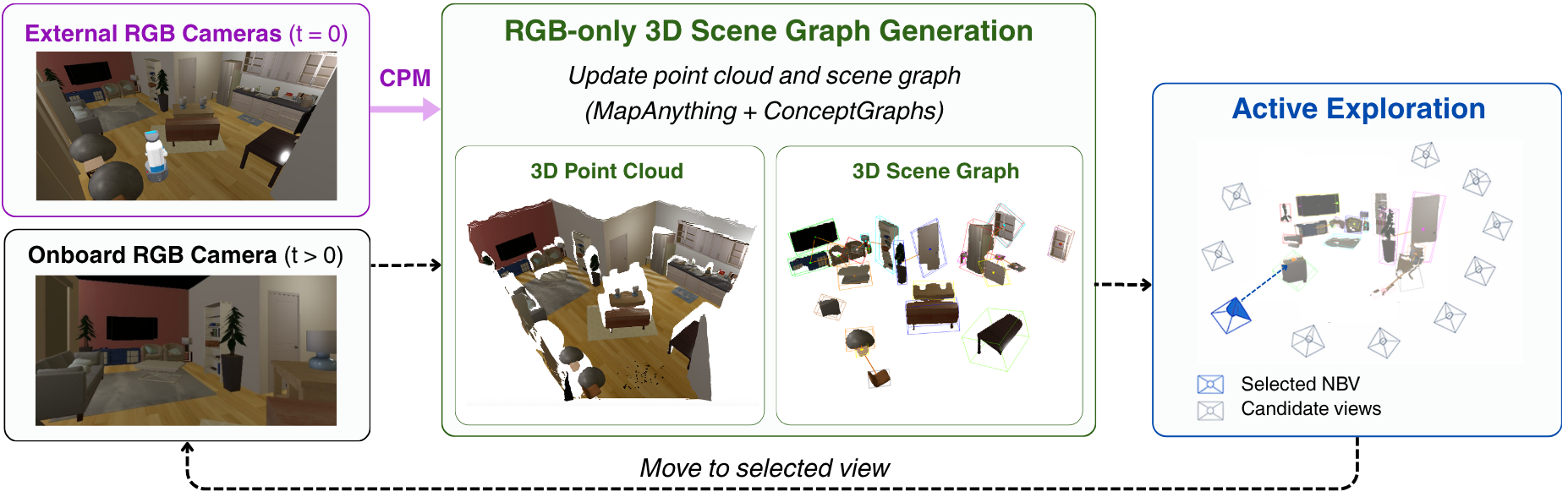}
\caption{System Overview. RGB observations from fixed external cameras serve as a Common Prior Map (CPM) and are processed by the RGB-only 3DSG generation pipeline to initialize the point cloud and 3D scene graph. For $t>0$, the onboard robot camera provides new RGB observations that incrementally update the point cloud and scene graph. The active exploration module then uses the current scene graph to select the Next-Best-View (NBV) and the robot moves to the selected viewpoint to continue exploration.}
\label{fig:overview}
\end{figure*}

The architecture is modular, separating 3D metric reconstruction, scene graph generation, and active exploration into distinct stages.

\subsection{System Overview}
The system operates as an incremental perception--action loop, illustrated in Fig.~\ref{fig:overview}. At initialization ($t=0$), RGB observations from one or more fixed external cameras serve as a \emph{Common Prior Map} (CPM). These observations are processed by the RGB-only 3DSG generation pipeline to initialize a point cloud and scene graph, providing semantic and geometric structure before robot motion begins.

For subsequent steps ($t>0$), the robot acquires an RGB image from its current viewpoint $v_t \in \mathcal{V}$, where $\mathcal{V}$ denotes the set of navigable viewpoints. The same RGB-only pipeline processes this observation to update the reconstructed point cloud $\mathcal{P}_t$ and the scene graph $\mathcal{G}_t $. Based on the current graph, the active exploration module selects the Next-Best-View (NBV) $v_{t+1}$ by maximizing expected information gain. The robot then moves to the selected viewpoint, acquires a new observation from the onboard camera, and repeats the loop, progressively extending and refining the initial CPM.

\subsection{RGB-Only 3DSG Generation}

\textbf{Depths and Poses Estimation.}
The pipeline leverages MapAnything~\cite{keetha2025mapanything} to infer scene geometry from a set of RGB images.
Given $N$ inputs, the model predicts in a single forward pass a \emph{factored} representation: per-view pixel ray directions $R_i$, up-to-scale depth maps $\tilde{D}_i$, poses $\tilde{P}_i$ in the reference frame of the first image, and a global metric scale factor $m \in \mathbb{R}$.
These outputs are composed to back-project 2D pixels into a metrically scaled 3D point cloud.
Since all poses are expressed in the reference frame of the first input image, fixing this reference - e.g., a known external camera location - is sufficient to anchor all subsequent observations in a consistent global frame, supporting reliable robot localization within the CPM's coordinate system.

\textbf{Open-Vocabulary Scene Graph Construction.}
RGB images, together with estimated depths and poses, are passed to ConceptGraphs~\cite{gu2024conceptgraphs}: SAM~\cite{kirillov2023segment} extracts class-agnostic instance masks, CLIP~\cite{radford2021clip} embeds each region into a semantic descriptor, the masks are projected into 3D using the MapAnything-derived depth and poses, and multi-view association incrementally merges detections into 3-D object nodes $o_j \in \mathcal{O}_t$. While the original ConceptGraphs pipeline queries proprietary LLMs to infer relationships between objects~\cite{gu2024conceptgraphs}, here spatial edges are instead derived through a deterministic, geometry-only procedure over oriented 3-D bounding boxes of the object nodes: for each ordered object pair, a fixed-priority set of predicates is evaluated - \textit{on top of / supported by} (vertical contact with footprint overlap), \textit{under / over} (vertical adjacency), \textit{inside} (volumetric containment), and \textit{next to} (horizontal proximity at similar height) - and at most one edge per ordered pair is assigned, yielding a reproducible scene graph. All geometric relation thresholds (e.g., maximum vertical gap, minimum footprint overlap, and horizontal proximity) are fixed across experiments and tuned empirically once per dataset.

\subsection{Active Semantic Perception}
The framework incorporates Active Semantic Perception (ASP)~\cite{tang2025active} to actively explore the unknown environment using the current 3DSG. At each exploration step, ASP reasons over the scene graph produced so far by the RGB-only pipeline and uses it as the structured state for action selection. An LLM samples plausible completions of the unobserved scene conditioned on the current graph, enabling the system to infer what may lie beyond the currently explored regions. For each candidate viewpoint $x$, ASP computes expected information gain as the mutual information between the predicted observation $Y_{k+1}$ and the current graph $G_k$:
\begin{equation}
I(Y_{k+1}; G_k \mid x) = H(Y_{k+1} \mid x) - H(Y_{k+1} \mid x, G_k)
\end{equation}
where $H(\cdot)$ denotes Shannon entropy. By selecting the viewpoint that maximizes information gain, ASP prioritizes locations that are expected to resolve semantic ambiguities in the partially observed scene (e.g., behind a door leading to a kitchen). In this setting, the 3DSG serves not only as the target representation, but also as the structured model that guides exploration. 

\subsection{External Cameras as Common Prior Maps}
Because the perception framework is strictly visual, observations from external RGB cameras can be processed by MapAnything identically to onboard egocentric images. In this work, we treat such observations as CPMs: a source of semantic and geometric prior information available before robot deployment. When used to initialize the system, this CPM provides partial semantic and geometric structure before robot motion begins, allowing ASP to exploit richer initial context and select more informative viewpoints from the earliest stages of exploration.

\section{Experiments and Results}
\label{sec:experiments}

We evaluate the proposed method in three experimental settings: static pipeline validation, active exploration, and multi-view external camera initialization. Across all settings, node quality is assessed automatically using joint semantic and geometric matching: predicted and ground-truth labels are embedded with a SentenceTransformer~\cite{reimers2019sentencebert}, and matches are accepted only when both semantic similarity and 3D localization constraints satisfy fixed thresholds. Quantitative evaluation is restricted to nodes, since datasets used in this work do not provide ground-truth relational annotations, while edges are generated by a deterministic geometry-based module that was extensively tested during development.

\subsection{Static RGB-Only 3DSG Pipeline Validation}
We first validate the core RGB-only 3DSG generation pipeline in a static setting, independently of both external camera initialization and active exploration. Specifically, we compare the original ConceptGraphs (CG), which uses depth observations and camera poses, with our RGB-only variant (CG-RGB), which replaces them with MapAnything predictions. The goal is to isolate the effect of this substitution and verify that the resulting RGB-only pipeline preserves scene-graph node quality. Evaluation was conducted on the Replica dataset~\cite{straub2019replica}, and Table~\ref{tab:static_eval} reports the results.

\begin{table}[ht]
\centering
\caption{Quantitative evaluation of scene graph node accuracy across seven Replica scenes (\texttt{room0-2}, \texttt{office0-3}). CG is the original ConceptGraphs, CG-RGB is the proposed RGB-only variant, with predicted depths and poses. Values are reported as mean $\pm$ standard deviation across scenes.1}
\setlength{\tabcolsep}{5pt}
\begin{tabular}{lccc}
\toprule
Experiment & Precision $\uparrow$ & Recall $\uparrow$ & F1-score $\uparrow$ \\
\midrule
CG     & $\mathbf{0.686 \pm 0.05}$ & $0.401 \pm 0.10$        & $0.499 \pm 0.08$        \\
CG-RGB (ours) & $0.615 \pm 0.07$        & $\mathbf{0.436 \pm 0.12}$ & $\mathbf{0.500 \pm 0.08}$ \\
\bottomrule
\end{tabular}
\label{tab:static_eval}
\end{table}

The results show strong parity between the two variants, with nearly identical F1 scores. MapAnything's noisier depth predictions introduce minor geometric inconsistencies, slightly lowering precision but preventing ConceptGraphs from overly aggressive merging of closely situated objects, thereby improving recall. Overall, these findings indicate that replacing ground-truth depth and pose with MapAnything estimates preserves the quality of the scene graph.

\subsection{Active Exploration with CPM Initialization}

\begin{figure*}[t]
\centering
\begin{minipage}{0.48\textwidth}
\centering
\includegraphics[width=\linewidth]{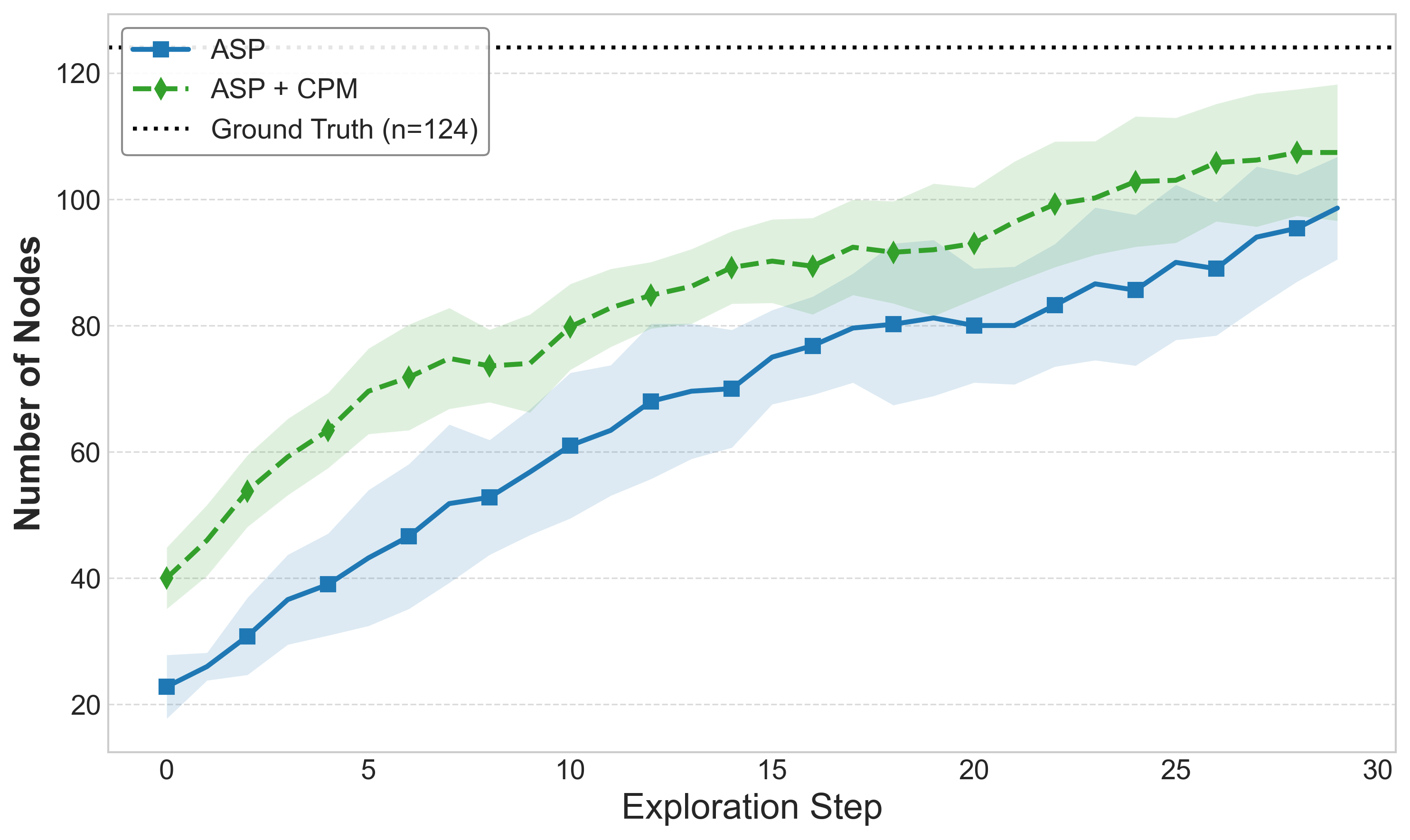}
\end{minipage}
\hfill
\begin{minipage}{0.48\textwidth}
\centering
\includegraphics[width=\linewidth]{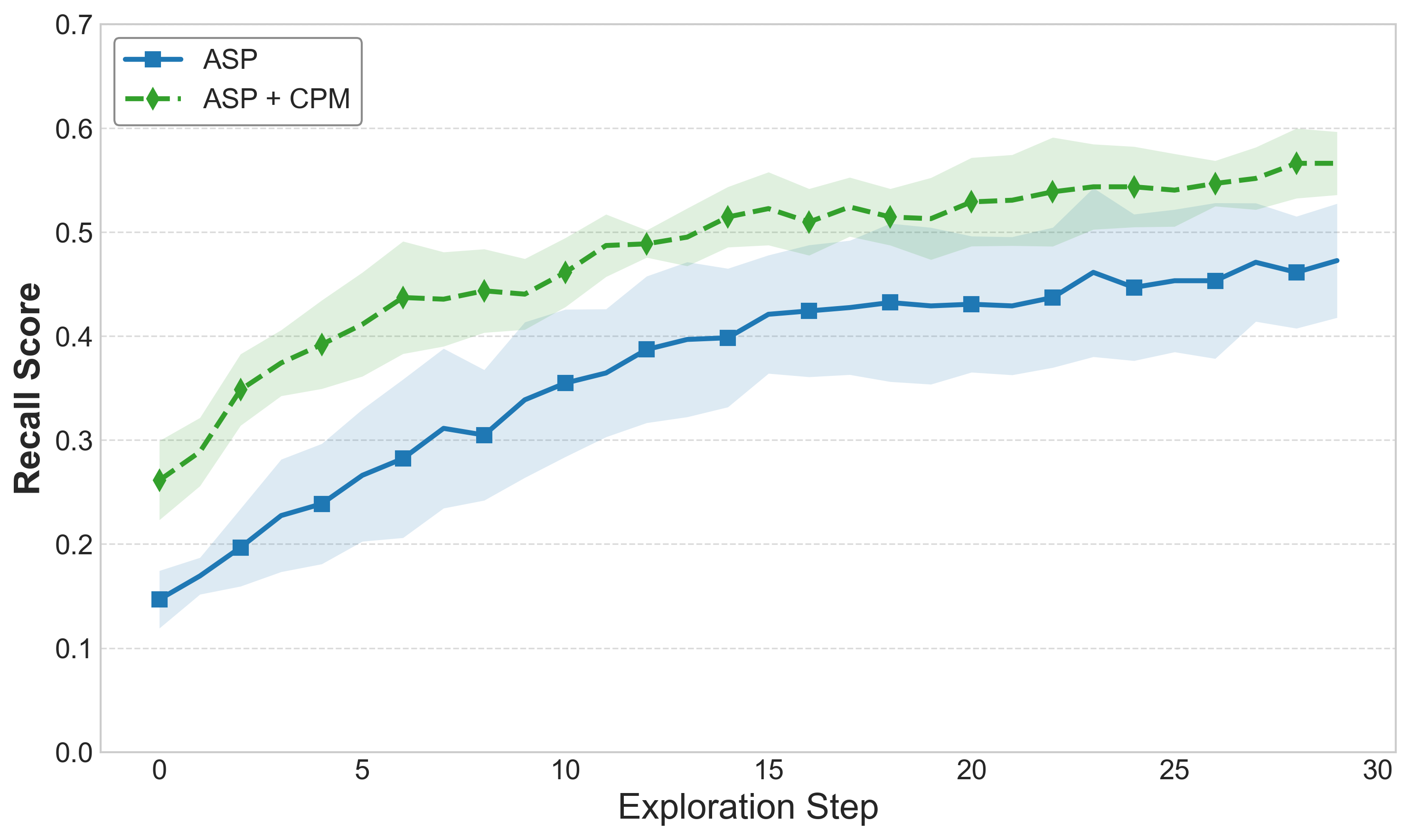}
\end{minipage}
\caption{Evolution of number of predicted nodes (left) and recall (right) for ASP, with and without CPM initialization from a single external camera (+CPM), over 30 exploration steps in \texttt{apartment 4}. Shaded areas denote standard deviation over different initial robot poses. The prior provides a strong initialization and sustains improved performance throughout exploration.}
\label{fig:active_expl}
\end{figure*}

The RGB-only active exploration pipeline reported in Fig.~\ref{fig:overview} was evaluated within the Habitat simulator \cite{savva2019habitat} using the six \texttt{apartment} scenes from the ReplicaCAD dataset~\cite{szot2021habitat2}, where local navigation to selected viewpoints is handled via Habitat’s built-in navigation utilities. \texttt{gemini-2.5-pro}~\cite{team2023gemini} was used as LLM for ASP. Experiments were run from 10 distinct starting positions over 30 exploration steps. We considered two settings: ASP operating from onboard robot observations only, and ASP initialized with one additional fixed external camera observation acting as a CPM prior. Since performance trends were consistent across the six apartments, the results reported in Fig.~\ref{fig:active_expl} focus on \texttt{apartment 4} as a representative scene.

Operating exclusively on onboard observations, ASP identified approximately 98 object nodes by step 30, corresponding to a recall of about 0.47. When initialized with a single overhead external camera, ASP started from a substantially richer scene graph, increasing the initial node count from approximately 23 to 40 (about +74\%) and the initial recall from about 0.15 to 0.26 (about +79\%), and maintained this advantage throughout exploration. By step 30, the CPM-initialized setting reached approximately 107 nodes and a recall of about 0.56, with the trajectory remaining consistently above the onboard-only case in both node count and recall. These results indicate that even a \textit{single} external camera can provide a strong semantic and geometric prior, improving scene graph completeness and supporting more effective semantic exploration from the first step.

\subsection{External Camera Infrastructure as Standalone CPM}
To quantify the standalone value of camera-infrastructure CPMs, we evaluated the RGB-only 3DSG generation pipeline using only fixed external cameras (one to three), without a mobile robot or active exploration. In total, 270 experiments were conducted across 90 ReplicaCAD scenes. The scenes were split into complex \texttt{apartments}  (dense clutter, $\sim$123 objects, used also in the previous experiment) and simpler \texttt{furnished rooms} ($\sim$25 larger objects). Table~\ref{tab:external_camera_results} reports the results.

\begin{table}[ht]
\centering
\caption{Scene graph accuracy using one to three fixed external cameras (\#Cam.) as standalone CPMs on ReplicaCAD \texttt{apartments} (Apt.) and \texttt{furnished rooms} (Fur.) scenes. No mobile robot is used. Values are mean $\pm$ std across scenes.}
\begin{tabular}{lcccc}
\toprule
Scene & \#Cam. & Precision $\uparrow$ & Recall $\uparrow$ & F1-score $\uparrow$ \\
\midrule
& 1 & \textbf{0.770 $\pm$ 0.054} & 0.198 $\pm$ 0.030 & 0.315 $\pm$ 0.041 \\
Apt. & 2 & 0.741 $\pm$ 0.036 & 0.232 $\pm$ 0.030 & 0.353 $\pm$ 0.036 \\
& 3 & 0.698 $\pm$ 0.031 & \textbf{0.301 $\pm$ 0.029} & \textbf{0.421 $\pm$ 0.030} \\
\midrule
& 1 & \textbf{0.566 $\pm$ 0.087} & 0.398 $\pm$ 0.071 & 0.465 $\pm$ 0.073 \\
Fur. & 2 & 0.540 $\pm$ 0.077 & 0.487 $\pm$ 0.071 & 0.510 $\pm$ 0.067 \\
& 3 & 0.486 $\pm$ 0.060 & \textbf{0.555 $\pm$ 0.065} & \textbf{0.516 $\pm$ 0.055} \\
\bottomrule
\end{tabular}
\label{tab:external_camera_results}
\end{table}

While 3 static viewpoints cannot match the recall of a 30-step active robotic exploration, they can successfully establish the environment's dominant structure. In complex \texttt{apartments}, three cameras achieved a recall of 0.301, while in \texttt{furnished rooms}, the recall reached 0.555. Precision declines slightly as cameras are added due to overlapping multi-view duplicate fragments, but the net F1-score confirms that fixed external RGB cameras can function as a high-quality CPM for semantic-geometric scene initialization, even in the absence of any mobile robot.

\section{Conclusion}
\label{sec:conclusion}
This paper presented an RGB-only framework for active 3D scene graph construction that uses observations from fixed external cameras as Common Prior Maps. The proposed pipeline builds 3D scene graphs directly from external RGB views, including already deployed surveillance cameras, and uses them as priors for downstream exploration and planning without additional hardware cost. In this way, it advances scene understanding by providing a structured semantic and geometric representation of the environment that can be incrementally refined through subsequent robot observations. In addition, because these cameras provide stable wide-field views, they can also improve the robustness of localization and mapping by anchoring the reconstruction in a consistent global reference frame and by offering broader spatial context across rooms. Experiments show that even a single external camera delivers substantial gains in initial scene coverage and improves the efficiency and quality of subsequent active exploration. Overall, the proposed approach enables robotic systems to leverage widely available camera infrastructure as a powerful source of geometric and semantic priors, opening practical new directions for scalable, infrastructure-assisted robot autonomy.

\bibliographystyle{IEEEtran}
\bibliography{biblio}

@article{ni2023deep,
  title={Deep learning-based scene understanding for autonomous robots: A survey},
  author={Ni, Jianjun and Chen, Yan and Tang, Guangyi and Shi, Jiamei and Cao, Weidong and Shi, Pengfei},
  journal={Intelligence \& Robotics},
  volume={3},
  number={3},
  pages={374--401},
  year={2023},
  publisher={OAE Publishing Inc.}
}

@article{team2023gemini,
  title={Gemini: a family of highly capable multimodal models},
  author={Team, Gemini and Anil, Rohan and Borgeaud, Sebastian and Alayrac, Jean-Baptiste and Yu, Jiahui and Soricut, Radu and Schalkwyk, Johan and Dai, Andrew M and Hauth, Anja and Millican, Katie and others},
  journal={arXiv preprint arXiv:2312.11805},
  year={2023}
}

@inproceedings{gu2024conceptgraphs,
  title={Conceptgraphs: Open-vocabulary 3d scene graphs for perception and planning},
  author={Gu, Qiao and Kuwajerwala, Ali and Morin, Sacha and Jatavallabhula, Krishna Murthy and Sen, Bipasha and Agarwal, Aditya and Rivera, Corban and Paul, William and Ellis, Kirsty and Chellappa, Rama and others},
  booktitle={2024 IEEE International Conference on Robotics and Automation (ICRA)},
  pages={5021--5028},
  year={2024},
  organization={IEEE}
}

@inproceedings{savva2019habitat,
  title={Habitat: A platform for embodied ai research},
  author={Savva, Manolis and Kadian, Abhishek and Maksymets, Oleksandr and Zhao, Yili and Wijmans, Erik and Jain, Bhavana and Straub, Julian and Liu, Jia and Koltun, Vladlen and Malik, Jitendra and others},
  booktitle={Proceedings of the IEEE/CVF international conference on computer vision},
  pages={9339--9347},
  year={2019}
}

@article{szot2021habitat2,
  title={Habitat 2.0: Training home assistants to rearrange their habitat},
  author={Szot, Andrew and Clegg, Alexander and Undersander, Eric and Wijmans, Erik and Zhao, Yili and Turner, John and Maestre, Noah and Mukadam, Mustafa and Chaplot, Devendra Singh and Maksymets, Oleksandr and others},
  journal={Advances in neural information processing systems},
  volume={34},
  pages={251--266},
  year={2021}
}

@article{keetha2025mapanything,
  title={Mapanything: Universal feed-forward metric 3d reconstruction},
  author={Keetha, Nikhil and M{\"u}ller, Norman and Sch{\"o}nberger, Johannes and Porzi, Lorenzo and Zhang, Yuchen and Fischer, Tobias and Knapitsch, Arno and Zauss, Duncan and Weber, Ethan and Antunes, Nelson and others},
  journal={arXiv preprint arXiv:2509.13414},
  year={2025}
}

@inproceedings{radford2021clip,
  title={Learning transferable visual models from natural language supervision},
  author={Radford, Alec and Kim, Jong Wook and Hallacy, Chris and Ramesh, Aditya and Goh, Gabriel and Agarwal, Sandhini and Sastry, Girish and Askell, Amanda and Mishkin, Pamela and Clark, Jack and others},
  booktitle={International conference on machine learning},
  pages={8748--8763},
  year={2021},
  organization={PmLR}
}

@article{straub2019replica,
  title={The replica dataset: A digital replica of indoor spaces},
  author={Straub, Julian and Whelan, Thomas and Ma, Lingni and Chen, Yufan and Wijmans, Erik and Green, Simon and Engel, Jakob J and Mur-Artal, Raul and Ren, Carl and Verma, Shobhit and others},
  journal={arXiv preprint arXiv:1906.05797},
  year={2019}
}

@inproceedings{yamauchi1997frontier,
  title={A frontier-based approach for autonomous exploration},
  author={Yamauchi, Brian},
  booktitle={Proceedings 1997 IEEE International Symposium on Computational Intelligence in Robotics and Automation CIRA'97.'Towards New Computational Principles for Robotics and Automation'},
  pages={146--151},
  year={1997},
  organization={IEEE}
}

@inproceedings{xu2017iterative,
  title={Scene graph generation by iterative message passing},
  author={Xu, Danfei and Zhu, Yuke and Choy, Christopher B and Fei-Fei, Li},
  booktitle={Proceedings of the IEEE conference on computer vision and pattern recognition},
  pages={5410--5419},
  year={2017}
}

@article{Li2024survey,
  title={Scene graph generation: A comprehensive survey},
  author={Li, Hongsheng and Zhu, Guangming and Zhang, Liang and Jiang, Youliang and Dang, Yixuan and Hou, Haoran and Shen, Peiyi and Zhao, Xia and Shah, Syed Afaq Ali and Bennamoun, Mohammed},
  journal={Neurocomputing},
  volume={566},
  pages={127052},
  year={2024},
  publisher={Elsevier}
}

@inproceedings{bae2023survey3dscenegraphs,
  title={A survey on 3D scene graphs: Definition, generation and application},
  author={Bae, Jaewon and Shin, Dongmin and Ko, Kangbeen and Lee, Juchan and Kim, Ue-Hwan},
  booktitle={International Conference on Robot Intelligence Technology and Applications},
  pages={136--147},
  year={2022},
  organization={Springer}
}

@article{catalano20253d,
  title={3D scene graphs in robotics: A unified representation bridging geometry, semantics, and action},
  author={Catalano, Iacopo and Zumaya, Carlos Cueto and Placed, Julio A and Civera, Javier and Bessa, Wallace Moreira and Pe{\~n}a-Queralta, Jorge},
  journal={Authorea Preprints},
  year={2025},
  publisher={Authorea}
}

@article{hughes2022hydra,
  title={Hydra: A real-time spatial perception system for 3D scene graph construction and optimization},
  author={Hughes, Nathan and Chang, Yun and Carlone, Luca},
  journal={arXiv preprint arXiv:2201.13360},
  year={2022}
}

@inproceedings{wu2021scenegraphfusion,
  title={Scenegraphfusion: Incremental 3d scene graph prediction from rgb-d sequences},
  author={Wu, Shun-Cheng and Wald, Johanna and Tateno, Keisuke and Navab, Nassir and Tombari, Federico},
  booktitle={Proceedings of the IEEE/CVF Conference on Computer Vision and Pattern Recognition},
  pages={7515--7525},
  year={2021}
}

@inproceedings{werby2024hierarchical,
  title={Hierarchical open-vocabulary 3d scene graphs for language-grounded robot navigation},
  author={Werby, Abdelrhman and Huang, Chenguang and B{\"u}chner, Martin and Valada, Abhinav and Burgard, Wolfram},
  booktitle={First Workshop on Vision-Language Models for Navigation and Manipulation at ICRA 2024},
  year={2024}
}

@article{rana2023sayplan,
  title={Sayplan: Grounding large language models using 3d scene graphs for scalable robot task planning},
  author={Rana, Krishan and Haviland, Jesse and Garg, Sourav and Abou-Chakra, Jad and Reid, Ian and Suenderhauf, Niko},
  journal={arXiv preprint arXiv:2307.06135},
  year={2023}
}

@article{jiang2024roboexp,
  title={Roboexp: Action-conditioned scene graph via interactive exploration for robotic manipulation},
  author={Jiang, Hanxiao and Huang, Binghao and Wu, Ruihai and Li, Zhuoran and Garg, Shubham and Nayyeri, Hooshang and Wang, Shenlong and Li, Yunzhu},
  journal={arXiv preprint arXiv:2402.15487},
  year={2024}
}

@inproceedings{kirillov2023segment,
  title={Segment anything},
  author={Kirillov, Alexander and Mintun, Eric and Ravi, Nikhila and Mao, Hanzi and Rolland, Chloe and Gustafson, Laura and Xiao, Tete and Whitehead, Spencer and Berg, Alexander C and Lo, Wan-Yen and others},
  booktitle={Proceedings of the IEEE/CVF international conference on computer vision},
  pages={4015--4026},
  year={2023}
}

@article{bajcsy2018revisiting,
  title={Revisiting active perception},
  author={Bajcsy, Ruzena and Aloimonos, Yiannis and Tsotsos, John K},
  journal={Autonomous Robots},
  volume={42},
  number={2},
  pages={177--196},
  year={2018},
  publisher={Springer}
}

@article{border2024surface,
  title={The surface edge explorer (see): A measurement-direct approach to next best view planning},
  author={Border, Rowan and Gammell, Jonathan D},
  journal={The International Journal of Robotics Research},
  volume={43},
  number={10},
  pages={1506--1532},
  year={2024},
  publisher={SAGE Publications Sage UK: London, England}
}

@article{tang2025active,
  title={Active Semantic Perception},
  author={Tang, Huayi and Chaudhari, Pratik},
  journal={arXiv preprint arXiv:2510.05430},
  year={2025}
}

@article{chen2025understanding,
  title={Understanding while Exploring: Semantics-driven Active Mapping},
  author={Chen, Liyan and Zhan, Huangying and Yin, Hairong and Xu, Yi and Mordohai, Philippos},
  journal={arXiv preprint arXiv:2506.00225},
  year={2025}
}

@article{ding2025sea,
  title={SEA: Semantic Map Prediction for Active Exploration of Uncertain Areas},
  author={Ding, Hongyu and Liang, Xinyue and Fang, Yudong and Wu, You and Shi, Jieqi and Huo, Jing and Li, Wenbin and Wu, Jing and Lai, Yu-Kun and Gao, Yang},
  journal={arXiv preprint arXiv:2510.19766},
  year={2025}
}

@article{robinson2026robot,
  title={Robot-relay: Building-wide, calibration-less visual servoing with learned sensor handover networks},
  author={Robinson, Luke and Gadd, Matthew and Newman, Paul and Martini, Daniele De},
  journal={Autonomous Robots},
  volume={50},
  number={1},
  pages={3},
  year={2026},
  publisher={Springer}
}

@article{buoso2025select2plan,
  title={Select2plan: Training-free icl-based planning through vqa and memory retrieval},
  author={Buoso, Davide and Robinson, Luke and Averta, Giuseppe and Torr, Philip and Franzmeyer, Tim and De Martini, Daniele},
  journal={IEEE Robotics and Automation Letters},
  year={2025},
  publisher={IEEE}
}

@inproceedings{reimers2019sentencebert,
  title={Sentence-bert: Sentence embeddings using siamese bert-networks},
  author={Reimers, Nils and Gurevych, Iryna},
  booktitle={Proceedings of the 2019 conference on empirical methods in natural language processing and the 9th international joint conference on natural language processing (EMNLP-IJCNLP)},
  pages={3982--3992},
  year={2019}
}

\end{document}